%% file: main.tex
\documentclass{article}
\usepackage{spconf}
\ninept % can comment this out to use 10pt instead.

\PassOptionsToPackage{hyphens}{url}\usepackage{hyperref}

\usepackage{url}

\usepackage{amsmath}
\usepackage{graphicx}
\usepackage{amssymb}
\usepackage{multirow}
\usepackage{hhline}
\usepackage{xcolor}
\usepackage{subfigure}
\usepackage{booktabs}

\usepackage{lipsum}
\usepackage{overpic}

\usepackage{pifont}% http://ctan.org/pkg/pifont

\usepackage{amsthm}

\newtheorem{proposition}{Proposition}
\newtheorem{definition}{Definition}
\newtheorem{conjecture}{Conjecture}
\newcommand{\vc}[1]{\ensuremath{\boldsymbol{#1}}}
\newcommand{\dofangle}{\text{DOF} }

\newcommand{\inR}[1]{\ensuremath{\in \mathbb{R}^{#1}}}
\newcommand{\rowvec}[2]{\ensuremath{\begin{bmatrix}#1 & #2\end{bmatrix}}}

\newcommand{\norm}[1]{\left\lVert#1\right\rVert}
\DeclareMathOperator*{\argmin}{arg\,min}

\usepackage{siunitx}

\title{Realizability of planar point embeddings from angle measurements}

\name{Frederike D{\"u}mbgen \quad Majed El Helou \quad Adam Scholefield}
\address{School of Computer and Communication Sciences, EPFL, Switzerland.}

\begin{document}
%~~% For small text font %~~%
% \small
\maketitle

\begin{abstract}
Localization of a set of nodes is an important and a thoroughly researched problem in robotics and sensor networks. This paper is concerned with the theory of localization from inner-angle measurements. We focus on the challenging case where no anchor locations are known. 

Inspired by Euclidean distance matrices, we investigate when a set of inner angles corresponds to a realizable point set. In particular, we find linear and non-linear constraints that are provably necessary, and we conjecture also sufficient for characterizing realizable angle sets. We confirm this in extensive numerical simulations, and we illustrate the use of these constraints for denoising angle measurements along with the reconstruction of a valid point set.
\end{abstract}


\input{1_intro_v2.tex}
\input{2_related_work.tex}
\input{2a_problem_setup.tex}
\input{3_dof.tex}
\input{4_constraints.tex}
\input{5_algorithms.tex}
\input{6_results.tex}
\input{7_conclusion.tex}

\vfill\pagebreak

\newpage

\small
\bibliographystyle{IEEEbib}
\bibliography{references_old,references_fd}

\end{document}

%% file: 1_intro_v2.tex
\section{Introduction} \label{sec:introduction}

Euclidean distance matrices (EDMs) are built from pairwise distances between points in Euclidean space~\cite{Gower1982, EDM2015, Dattoro2015}. They have a well-defined structure and properties, such as a rank constraint, which allows partial and noisy EDMs to be completed and denoised. This has been exploited in numerous applications including psychometrics, machine learning and room geometry reconstruction, to name a few. Since EDMs consider pairwise distances between arbitrary points, they correspond to localization problems of multiple nodes without anchors. These problems occur in various static and dynamic settings, for instance when the nodes are sensor networks \cite{Chowdhury2016},  collaborating drone swarms~\cite{Guerra2019} or firefighters~\cite{Nilsson2014}.

In many applications, angle measurements can be exploited, complementing or replacing distances. For example, in localization, distances are estimated from received signal strength (RSS) or time of flight (TOF), whereas angles can be estimated from phase differences in multi-antenna systems. Angles have the advantage of removing the necessity of synchronization between nodes, and degrade less with distance than for instance RSS. Angle of arrival (AoA) will appear in the next Bluetooth standard \cite{Bluetooth} and will be an interesting alternative for Internet of Things (IoT) applications. 

For inner-angle measurements, no exhaustive theory exists that can be exploited to denoise and complete angle measurements in a similar way. In this paper, we attempt to characterize inner angles between points in Euclidean space. In particular, we are interested in the question of realizability, which tells us if we can find a point set in a given dimension that exactly satisfies a given set of angle measurements. 

This paper makes three main contributions. 

\noindent
\textbf{a)} We find necessary conditions for inner-angle measurements to be realizable, consisting of linear and non-linear constraints which we exactly characterize. 

\noindent
\textbf{b)} We observe that the proposed conditions are also sufficient in extensive simulations.

\noindent
\textbf{c)} By exploiting the above conditions, we construct an angle denoising and point recovery algorithm. Using this algorithm, we characterize the noise regimes in which angle measurements are preferred over distance measurements in simulation. 
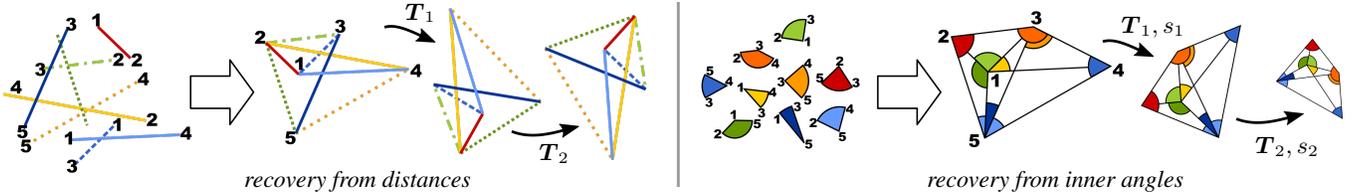
\begin{figure*}[ht]
    \centering
    %\begin{overpic}[width=\textwidth,grid,tics=2]{figures/overview.png}
    \begin{overpic}[width=\textwidth]{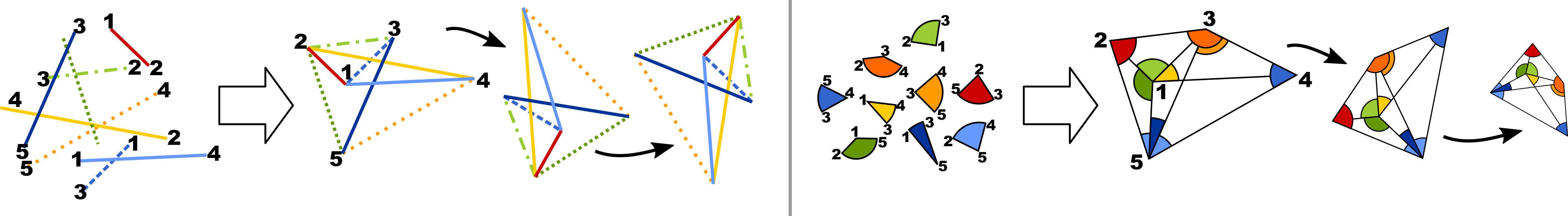}
    \put (18,0) {\textit{recovery from distances}}
    \put (69,0) {\textit{recovery from inner angles}}
    \put (30,12.5) {$\vc{T}_1$}
    \put (40,2) {$\vc{T}_2$}
    \put (83.5,11.5) {$\vc{T}_1, s_1$}
    \put (93.5,2.5) {$\vc{T}_2, s_2$}
    \end{overpic}
    
    \vspace{-0.5em}
    \caption{Recovering point embeddings from distances (left) vs. inner angles (right). Three possible reconstructions are shown for each setup. Inter-point distances are invariant to rigid transformations $\vc{T}_i$, whereas inner angles are invariant to both rigid transformations and scale $s_i$.}
    \label{fig:dof}
\end{figure*}

%% file: 2_related_work.tex
\section{Related Work} \label{sec:related_work}

%\textbf{Related Work} 
When measuring dissimilarities between points in terms of distances, a rich literature on distance geometry, and in particular EDMs, can be exploited for purposes such as denoising, completion, and point recovery \cite{EDM2015, Liberti2014}. In particular, it was shown that a matrix $\vc{D}\inR{N\times N}$ is an EDM if and only if $-\frac{1}{2}\vc{J} \vc{D}\vc{J}$
is positive semi-definite, where $\vc{J}\inR{N\times N}$ is the geometric centering matrix, and $N$ is the number of points \cite{Gower1982}. By definition, if \vc{D} is an EDM, there exists a point embedding in some dimension $d$ for which $D_{ij}=\norm{\vc{p_i}-\vc{p_j}}^2$, with $\vc{p_i},\vc{p_j}\inR{d}$. Thanks to this property, point coordinates can be recovered from complete pairwise distances through a simple SVD, or via semi-definite programs (SDPs) for the incomplete case. The question we ask in this paper is if we can provide a similar characterization for angles: given a set of angle measurements, can these angles be realized by a point set?  

Various methods have been proposed that incorporate both distance and angle information in point recovery algorithms \cite{Billinge2018, Macagnano2013, Baechler2018, Doherty2001, Biswas2005, Tomic2019}.
For instance, the concept of \textit{edge kernels} was introduced in \cite{Macagnano2013}, and the authors show that the angle and distance information can be estimated through matrix factorization. Although elegant, this concept does not address realizability. In fact, it was shown in \cite{Baechler2018} that the recovered angles cannot in general be realized, and additional linear constraints are added to the optimization problem to remedy this. 
Other methods have shown improved performance over distance-based SDPs by adding hyperplane-based constraints \cite{Doherty2001} or constraints based on the cosine law \cite{Biswas2005}. 
%\cite{Doherty2001} some convex constraints for localization, really well cited (2000) 
%\cite{Biswas2005} semidefinite relaxation of angle measurements? cotes Doherti2001 and says it is more exact. 

%\cite{Coluccia2019} has good arguments why node localization without anchors is becoming more and more important. Also does centralized approach. 
%\cite{Tomic2019} (this guys has hundreds of papers on the same topic (RSS + AoA)). 

%, and it is questionable whether the approach is optimal under common noise models, such as Gaussian noise. 
% in contrast, we find maximum likelihood estimate, and we can provide some conditions. 
Localization from angles only has also been studied thoroughly. In vision-based systems, bearing-only localization is solved through trilateration~\cite{Hartley2004}. In robot and sensor networks, distributed methods~\cite{Basu2006, Bruck2009, Yan2007}, Simultaneous Localization and Mapping (SLAM)-based solutions \cite{Deans2005, Lourenco2018}, and centralized approaches \cite{Guerra2019, Biswas2005} have been proposed. In the latter, a central processing unit collects data from all nodes and recovers their locations in a global manner. In distributed approaches, the nodes are self-localizing, and in most SLAM-based algorithms, the location of an active device and a (usually passive) map of landmarks are iteratively updated.   

%An important application area bearings-only localization is bearing-only SLAM \cite{Deans2005, Lourenco2018}, a subgroup of sparse visual SLAM (V-SLAM) in which no depth information of sparse features is available. Since angle measurements do not recover the notion of scale, inertial measurements are required to embed the constructed map in a global reference frame.

In this paper, we focus on the complete, centralized localization problem, meaning that we globally process all inner angles at once. We do not require knowledge of the node's orientations or any anchor nodes. In practice we rarely have access to all measurements, in which case the proposed theory can be used for completion and denoising. %Note however that this completeness is not a strong assumption, as the provided conditions can be used to fill in missing measurements. 
Our work is close in nature to
an early work on angle-based planar graph embedding \cite{Battista1996}. The authors prove conditions for an angle set to correspond to a planar and convex planar graph with no crossing edges. We relax these assumptions to graphs lying in 2 dimensions, with arbitrary edge layouts. By doing so, we can construct an optimization problem on angles only, which characterizes the maximum likelihood estimator under a Gaussian noise assumption, and ensures realizability.

%Although not directly addressed here, the proposed method recovers a unique embedding and is as such also related to graph rigidity theory \cite{Zelazo2014}. 

%\cite{Wang2018} maybe also incorporate our kinds of constraints? not quite clear. Anyways, they assume angle measurements without knowledge of node locations, so relative angle measurements. 
%\cite{Nollenburg2017} some conditions for greedy drawings of tree to be well-defined, probably related to us. 
%Computer vision stuff: seems too far fetched
%Graph embedding: \cite{Yan2007}, cited by \cite{Harandi2011}(CVPR). Grassmanian manifolds etc. do they also include some angular similarity measures?
%\cite{Paille2015} meshes using dihedral angle-based maps. computer graphics stuff. 

%% file: 2a_problem_setup.tex
\section{Preliminaries and Problem Setup} \label{sec:preliminaries}

\subsection{Problem setup}

We denote by $P=\left\{\vc{p_i}\inR{d} | i=1\ldots N\right\}$ the point embedding that we aim to recover, where $d$ is the (known) dimension of the embedding space and $N$ is the number of points. We focus on the two-dimensional case, but we use the general notation $d$ where applicable. 
The non-directed inner angle measured at point $\vc{p_i}$ between two points $\vc{p_j}$ and $\vc{p_k}$ is denoted by $\theta_i(j, k) = \theta_i(k, j) \in [0, \pi]$. As we only consider inner angles, we drop the ``inner'' attribute in what follows. We introduce the index set $\mathcal{I}$, which contains all triplets $(i, j, k)$ of angles, with $|\mathcal{I}|=N(N-1)(N-2)/2:=M$. By introducing an order on $\mathcal{I}$, denoted by $m(i,j,k)=1\ldots M$, we stack all angles in a vector denoted by $\vc{\theta} \inR{M}$. In this work we do not address the labeling problem, so we assume we know the indices $(i, j, k)$ for each $m$.

For the scope of this paper, we rule out degenerate point sets in which any angle $\theta_m$ is exactly either zero or $\pi$. When randomly choosing points, such setups occur with probability zero, and in practice they should be avoided because of ill-conditioning.  

We define the notion of a point set's \textit{equivalency class} as follows: Two point sets belong to the same equivalency class if each node has the the same circular clockwise sequence of edges in both point sets. The same notion was introduced in \cite{Battista1996}. As we will see later, this helps in setting up linear constraints since the linear constraints matrix \vc{A} is the same for all point sets in the same class. 

We conclude this section by defining the notion of a realizable angle vector, which is the main concept we study.

\begin{definition}\label{def:consistent}
    An angle vector $\vc{\theta}$ is realizable if and only if there exists a point embedding $P$ such that $\angle(\vc{p_j}-\vc{p_i}, \vc{p_k}-\vc{p_i}) = \theta_{m(i, j, k)}$ for all index triplets in $\mathcal{I}$.
\end{definition}

%% file: 3_dof.tex
\subsection{Degrees of freedom of realizable angles} \label{sec:dof}

%We lay down the foundations of degrees of freedom in angle-constrained point sets. 
When we measure the relative dissimilarities of points using distances or angles, some information is irrecoverably lost, as visualized in Figure \ref{fig:dof}. Using distances, the absolute translation ($d$ parameters), orientation and reflection  ($d(d-1)/2$ parameters) of the point set are lost. Although $N$ points in $d$-dimensional space have $Nd$ degrees of freedom, the pairwise differences only retain $Nd - d - d(d-1)/2 = Nd - d(d+1)/2$
degrees of freedom. When measuring only angles, we additionally lose scale information, leading to one degree of freedom less:
\begin{equation}
    \dofangle = Nd - \frac{d(d+1)}{2} - 1\text{.}
\end{equation}
\noindent We can conclude that we need at least \dofangle angle measurements to accurately recover a point embedding, up to a rigid transformation and scale --- the best that can be done from angle measurements.

\subsection{Recovering points from angles}\label{sec:recovery}

To recover point sets from the given angles $\vc{\theta}$, we apply the following standard build-up algorithm (similar algorithms have been studied using distances \cite{Alencar2019} or angles \cite{Billinge2018}). We start by fixing the first two points (\vc{p_1}, \vc{p_2}) arbitrarily. The subsequent points $\vc{p_n}$, $n=3\ldots d+1$ can be found using $n-1$ angles from the previous points. Finally, all remaining points are uniquely defined using $d+1$ angle measurements.  

Note that the minimum number of measurements required in the described algorithm is
\begin{equation} \label{eq:num_required}
    \sum_{i=2}^{d}i + (N - d + 1)(d + 1) = N(d+1) - \frac{d(d+1)}{2} + d,
\end{equation}
which is exactly $N+d+1$ more than the lower bound \dofangle. %In fact, we can only recover degenerate point configurations with the lower bound of number of measurements.

We stress that, given a full angle vector \vc{\theta}, this build-up algorithm only uses a subset of the angles for reconstruction. Unless the angles are realizable, the reconstruction accuracy thus depends on the (arbitrary) choice of angles used for reconstruction, and there might be a discrepancy between the non-used measured and reconstructed angles. In the next section, we provide constraints on the angles in vector \vc{\theta} that must be satisfied for them to correspond to a valid point set. These constraints can be used to find the closest realizable angle vector. By doing this denoising before applying the build-up algorithm, every selection of angles results in the same reconstruction (up to scale and rigid transformation).

%We conclude this section by formally stating the necessary condition for unique reconstruction.
%\begin{definition}
%   A point embedding $P$ from a consistent angle set $\Theta$ is unique if every other point embedding satisfying $\Theta$ is a rigid transformation or scaled version of $P$.
%\end{definition}

%\begin{proposition}\label{prop1}
%    An angle set $\Theta$ needs at least \dofangle measurements to allow for a unique point embedding.
%\end{proposition}

%Figure \ref{fig:reconstruction} shows a reconstruction example visualizing the origins of this equation. Note that if the point 4 lied on the line defined by points 1 and 2, only two angles would be sufficient for its recovery: $\theta_1(2, 4)=0$ and for example $\theta_3(1, 4)$. This is what we mean by "degenerate" point configurations in which less measurements are sufficient, and in the following we assume that such configurations occur with 0 probability and are thus neglegible.

%% file: 4_constraints.tex
\section{Characterization of realizable angles}\label{sec:constraints}

The intuition of our realizability conditions is the following: 
Since an angle vector has $M$ elements but only \dofangle degrees of freedom, $M-\dofangle$ of the measurements are redundant. %We can construct a point set using fewer angles, shown in \eqref{eq:num_required}, but we want to ensure that the redundant angles also match the recovered point set. 
In what follows, we construct a set of $L$ linear constraints and $K$ non-linear constraints on the angles, such that the $M-\dofangle$ residual degrees of freedom are correctly eliminated ($L+K=M-\dofangle$).

%As we show in Section \ref{sec:algorithms}, these realizability guarantees can be exploited for denoising angle measurements and for point recovery algorithms.
%$\vc{A}\vc{\theta}=\vc{b}$ and a set of $K$ non-linear constraints $f_i(\theta)=0$. As we will see later, $K+L=M-\dofangle$ and we can thus state

\subsection{Linear constraints}

We can impose two types of linear constraints on angles: the first type addresses angles measured at a single point, and the second type addresses angles in convex polygons. For all angles measured from one point, any subset of adjacent angles must sum up to form the bigger angle. In convex polygons of size $m\geq3$, the angles must sum up to $(m-2)\pi$. Numerous such constraints exist, but not all of them are linearly independent. For example, we can consider only triangles, as they are the building blocks of higher-order polygons. We provide a method that constructs exactly the maximum number of linearly-independent constraints, defined by $L = L_{single} + L_{triangle}$, with
\begin{align}\label{eq:linear}
    L_{single} = N\sum_{i=1}^{N-2}i, \text{ and }\,\,  
    L_{triangle} = \binom{N-1}{2}.
\end{align}

For the \textit{single} constraints at each node $n$, we define an order of the outgoing edges, $k=1\ldots N-1$, and the number of considered angles $\ell=2\ldots N-1$. Then, for each starting index $k=1\ldots N-1-\ell$ we impose that $\sum_{i=k}^{k+\ell}\theta_n(i, i+1)=\theta_n(k, k+\ell)$, or that $\sum_{i=k}^{k+\ell}\theta_n(i, i+1)=2\pi$. The order and condition can be determined through a combinatorial algorithm or from prior knowledge.
Moving to the \textit{triangle} constraints, one can find all constraints linearly independent from the \textit{single} constraints by choosing one single corner and imposing the constraint in all triangles involving that corner.
We represent the obtained linear constraints in matrix form $\vc{A}\vc{\theta}=\vc{b}$, with $\vc{A}\inR{L\times M}, \vc{b}\inR{L}$.

%We show in simulation that all other constraints (higher order polygon etc.), are linear combinations of the constraints from \eqref{eq:linear} in Section \ref{sec:results_linear}. 

\subsection{Automatic creation of linear constraints}

Constructing the constraints matrix $\vc{A}$, with the previously proposed combinatorial algorithm, can be time consuming. If we know the equivalency class of our point set, we can greatly speed up the process by learning the constraints matrix from simulations. If the equivalency class is not known a priori, it can be determined through a rough initial point estimate using the noisy or partially denoised angles.

We learn the constraints matrix by randomly generating at least $T \geq L$ point sets in the correct equivalency class, and reading off the angles $\vc{\theta}_{(t)}, t=1\ldots T$.  Since all points sets have the same order of edges per node, they share the same linear conditions and $\vc{A}\vc{\theta}_{(t)}=\vc{b}$, must be satisfied for each angle vector. Therefore we can write $\vc{\Theta}\vc{C}^\top=\vc{0}$, with
\begin{equation}
\vc{\Theta}=\rowvec{\vc{\theta}_{(t)}^\top}{ -1}_{t=1}^{T} 
%\inR{C\times (M+1)}
\text{, and } \vc{C} = \rowvec{\vc{A}}{\vc{b}}. %\inR{L\times(M+1)}
%   \begin{bmatrix}
%   \vc{\theta}_{(1)}^\top & -1 \\
%   \vc{\theta}_{(2)}^\top & -1 \\
%   ... \\
%   \vc{\theta}_{(K)}^\top & -1 \\
%   \end{bmatrix}
%   \begin{bmatrix}
%   \vc{A}^\top \\
%   \vc{b}^\top
%   \end{bmatrix} := \vc{\Theta} \vc{C}^\top
%   =\vc{0}_{}.
\end{equation}

In order to determine $\vc{C}^\top$, we can find a basis of the null space of the matrix $\vc{\Theta}$. A valid basis can be found by extracting the $L$ last columns of $\vc{V}$ from the SVD given by $\vc{C}^\top=\vc{U}\vc{\Sigma}\vc{V}^\top$.
As opposed to the combinatorally-obtained matrix \vc{A}, the learned matrix is not sparse. Since each constraint acts on all angles, this matrix leads to smoother denoising when incrementally adding constraints (see Section \ref{sec:results_realizability}).
%Figure \ref{fig:matrices} shows, for one chosen point set, the analytical and learned constraints matrices. 

\subsection{Non-linear constraints}

%We show in Figure \ref{fig:uniqueness} two different angle sets for which the linear constraints are satisfied, but only one of them is truly realizable by a point embedding. In fact there exists an infinite amount of non-realizable angles: we can add and subtract the same $\Delta$ from angles ... and ... without violating the linear constraints. To remedy this, we remove the residual degrees of freedom by adding $K$ non-linear constraints. 

As the linear constraints are not enough to eliminate the residual degrees of freedom, we need to resort to non-linear constraints. To this end, we can impose the sine law (or equivalently, the cosine law) in each triangle. In order to remove the distance parameters, we process two adjacent triangles together, which allows to eliminate the distances and obtain
\begin{align}
	f_k(\vc{\theta})=\frac{{\sin\theta_b(a, c)}}{{\sin\theta_b(a, d)}} 
\frac{{\sin\theta_c(a, d)}}{{\sin\theta_c(a, b)}} 
\frac{{\sin\theta_d(a, b)}}{{\sin\theta_d(a, c)}}-1=0,
\end{align}
where $a$ to $d$ denote the four corners of the $k$th quadrilateral. 
Repeating this for all quadrilaterals, we obtain a total of $K = \binom{N}{4}$ non-linear constraints.

%\subsection{Number of linear and non-linear constraints}\label{sec:results_linear}

\begin{figure}
    \centering
    \includegraphics[width=\linewidth, draft=false]{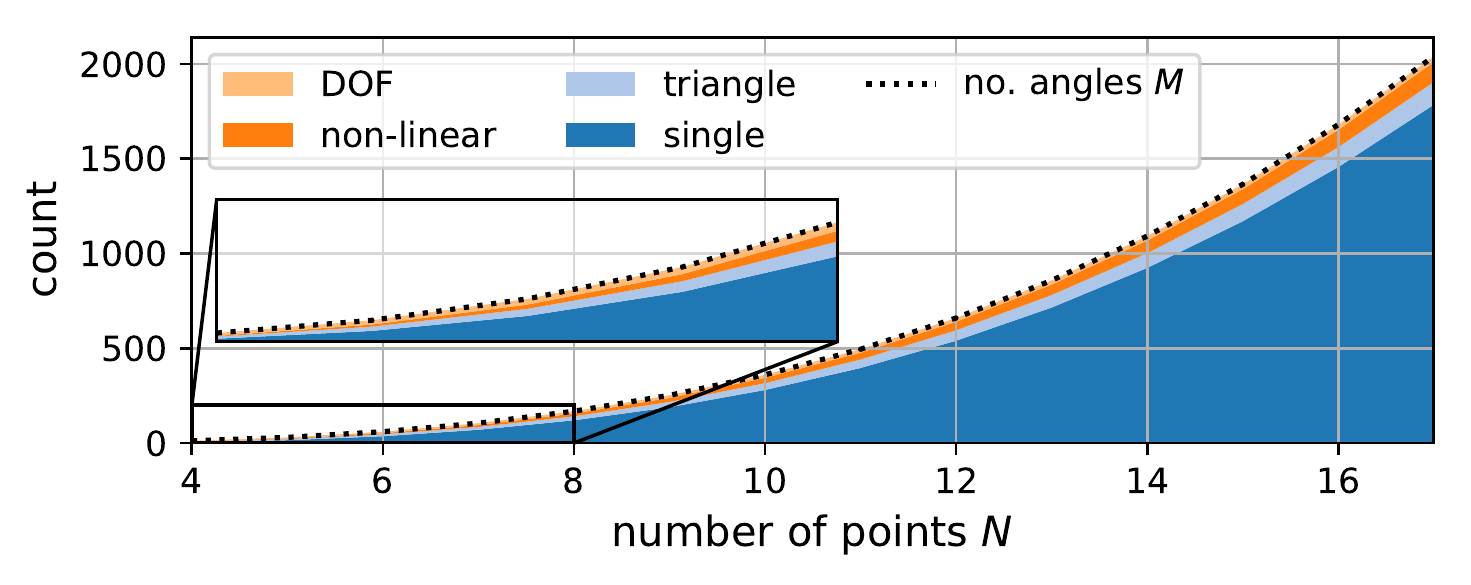}
    \vspace{-2.5em}
    \caption{Visualization of the number of angles $M$ as the sum of the degree of freedom (DOF) and the number of different constraint types, for increasing numbers of points $N$.}
    \label{fig:linear}
\end{figure}

\subsection{Guarantees}

Summing the number of linear and non-linear constraints found with the DOF of realizable angles gives the number of angles $M$, as desired:
\begin{equation}\label{eq:sum}
    M= DOF + L_{single} + L_{triangle} + K.
\end{equation}
%as shown in \eqref{eq:sum}.
%\begin{figure*}[h]
%\begin{equation}\label{eq:sum}
%    \frac{N(N-1)(N-2)}{2}=\underbrace{Nd-\frac{d(d+1)}{2} - 1}_{DOF} + \underbrace{N\frac{(N-2)(N-1)}{2}}_{L_{single}} + \underbrace{\binom{N-1}{2}}_{L_{triangle}} + \underbrace{\binom{N}{4}}_{K}
%\end{equation}
%\end{figure*}

Figure \ref{fig:linear} visualizes the different components of this equation. We can see that the majority of measurements can be constrained through linear constraints, so one might argue that the non-linear constraints are negligible. As we show in Section \ref{sec:results}, these constraints are however crucial for angle realizability, and for good reconstruction accuracy. 

We conclude by establishing necessary and sufficient guarantees for angle realizability. The proposed constraints are induced from well-known geometry laws, thus the following proposition holds: 

\begin{proposition}\label{necessary}
If \vc{\theta} is realizable, then both linear and non-linear constraints are satisfied: $\vc{A}\vc{\theta}=\vc{b}$ and $f_i(\vc{\theta})=0\,\text{ for }\,i=1\ldots K$.  
\end{proposition}

We have seen that by combining the proposed constraints, we get exactly $L+K=M-\dofangle$ constraints. We thus constrain all redundant measurements, and we can put forward the following conjecture.  

\begin{conjecture}\label{sufficient}
If both linear and non-linear constraints are satisfied, then the angle set \vc{\theta} is realizable. 
\end{conjecture}

We provide extensive simulation results that empirically verify this conjecture in Section \ref{sec:results_realizability}. If the conjecture is true, it follows that the constraints provided here are sufficient and necessary, as stated in the following result.

\begin{conjecture}\label{equivalent}
An angle set \vc{\theta} is realizable if and only if both linear and non-linear constraints are satisfied.
\end{conjecture}

%% file: 5_algorithms.tex
\section{denoising and recovery algorithm} \label{sec:algorithms}

 Possible applications of Conjecture \ref{equivalent} include outlier rejection (removing spurious measurements which do not satisfy the conditions), labeling (finding an order of angles such that the constraints are satisfied) and completion (filling in missing angle measurements such that they satisfy the constraints). 
For the purpose of this paper, we propose a denoising and point recovery algorithm.

Given a noisy angle vector $\vc{\widetilde{\theta}}$, we want to find a point embedding with angles as close to the measurements as possible. Thanks to Conjecture~\ref{equivalent} we can solve this in two steps: first, we find the maximum likelihood estimate (under zero-mean Gaussian noise assumption) of a realizable angle set by solving
\begin{equation}\label{eq:optimization}
    \begin{aligned}
    \vc{\hat{\theta}} = 
    & \argmin_{\vc{\theta} \inR{M}} || \vc{\widetilde{\theta}} - \vc{\theta}||\text{,} \\
    \text{such that } &\vc{A}\vc{\theta} = \vc{b} \text{, }
    f_k(\vc{\theta})=0 \text{ for }k=1\ldots K\text{.}
\end{aligned}
\end{equation}

The non-linear constraints render this problem non-convex, and no existing algorithm is guaranteed to find its optimum. As we see in Section \ref{sec:results}, standard solvers do however show good performance and convergence properties. %Although not absolutely necessary, we found that adding bounds on angles as $\theta_i\in (0, \pi) \text{ for } i=1\ldots M$ improved convergence speed.
Knowing that the resulting angle set $\vc{\hat{\theta}}$ is realizable, we use the simple build-up algorithm as described in Section \ref{sec:recovery} for point recovery.
If knowledge of some anchor points is provided, we align and scale the obtained point set to the anchor points using the orthogonal Procrustes transform \cite{Procrustes}.% We use this method in the next section to evaluate the reconstruction accuracy. 

%Introduce denoising algorithm, can we do more than simply denoising? How about completion? How about realizability check? How about finding best embedding dimension? Create simple algorithms that we can then evaluate in simulations in next section. 

%% file: 6_results.tex
\begin{figure}[t]
    \centering
    \setlength{\tabcolsep}{0pt}
    \begin{tabular}{cc}
    \textit{~~~~~~~~analytic constraints} & \textit{learned constraints} \\
    \includegraphics[height=5.4cm, draft=false]{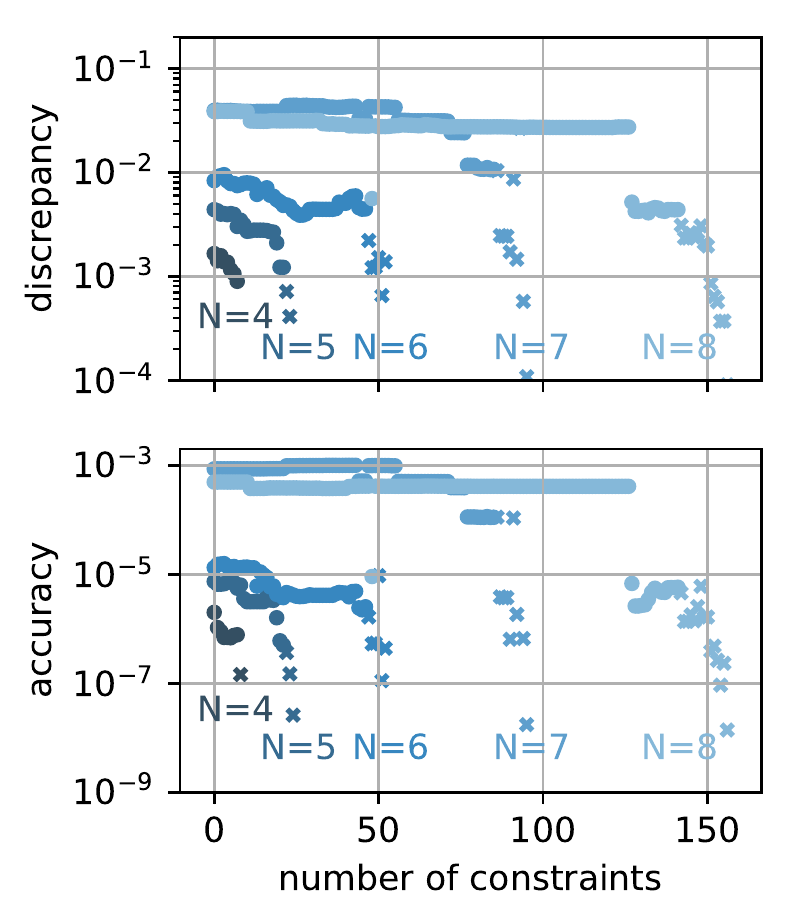}
    & \includegraphics[height=5.4cm, draft=false]{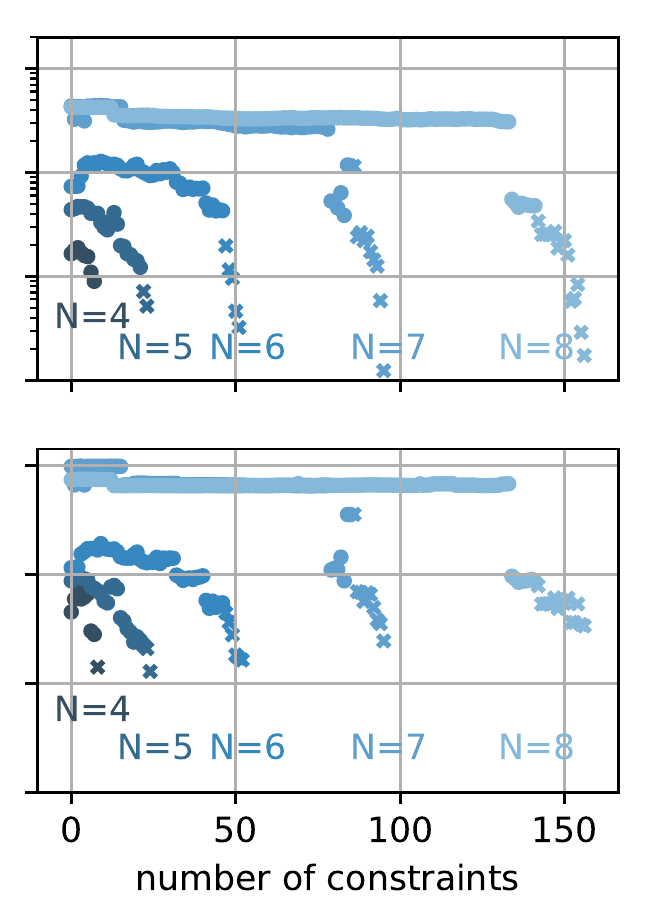}
    \end{tabular}
    \vspace{-1.2em}
    \caption{First row: discrepancy error vs. number of imposed constraints, for different numbers of points $N$, drawn uniformly from a unit square. Second row: reconstruction accuracy for the same setups. The angle noise is fixed at $\sigma_a=10^{-3}$.}
    \label{fig:discrepancy}
\end{figure}

\section{Simulation Results} \label{sec:results}

In this section, we validate the theory established in Section \ref{sec:constraints} through extensive numerical simulations and evaluate the algorithm's performance.\footnote{The code is available at \textit{github.com/duembgen/AngleRealizability}.}
We generate random point sets by uniformly picking points from a square of given side length. Simulated noise is always assumed zero-mean Gaussian. 
To avoid numerical issues we ignore point sets that have at least one angle smaller than \SI{1e-3}{rad} in absolute value.
We use the SLSQP solver provided by \texttt{scipy} \cite{scipy} to solve \eqref{eq:optimization}. 

\subsection{Realizability study}\label{sec:results_realizability}

To study the validity of Conjecture \ref{equivalent}, we design the following experiment: Given angles corrupted by zero-mean Gaussian noise of variance $\sigma_a^2$, we denoise the angles using \eqref{eq:optimization} and recover the point set using the build-up algorithm from Section \ref{sec:recovery}. From the obtained point set, we read off the angles. Clearly, the angles used for reconstruction equal the denoised angles. The unused angles, however,  only equal the denoised angles if the constraints in Conjecture \ref{equivalent} are sufficient. 
We report the results of this experiment in the top row of Figure~\ref{fig:discrepancy}.
We define the \textit{discrepancy} error as the MAE between the denoised and reconstructed angles, and calculate this error for 20 random realizations of each parameter set. We confirm that the discrepancy only approaches zero when all constraints are added. The learned constraints (right column) lead to a more gradual decrease than analytical constraints, because they are not sparse. Finally, we note that the difference in errors can be explained by the convergence to different local minima of \eqref{eq:optimization}. 
We conclude that the combination of the proposed constraints leads to realizable angles.

\subsection{Evaluation of proposed algorithm}

\begin{figure}[t]
    \centering
    \includegraphics[width=\linewidth, draft=false]{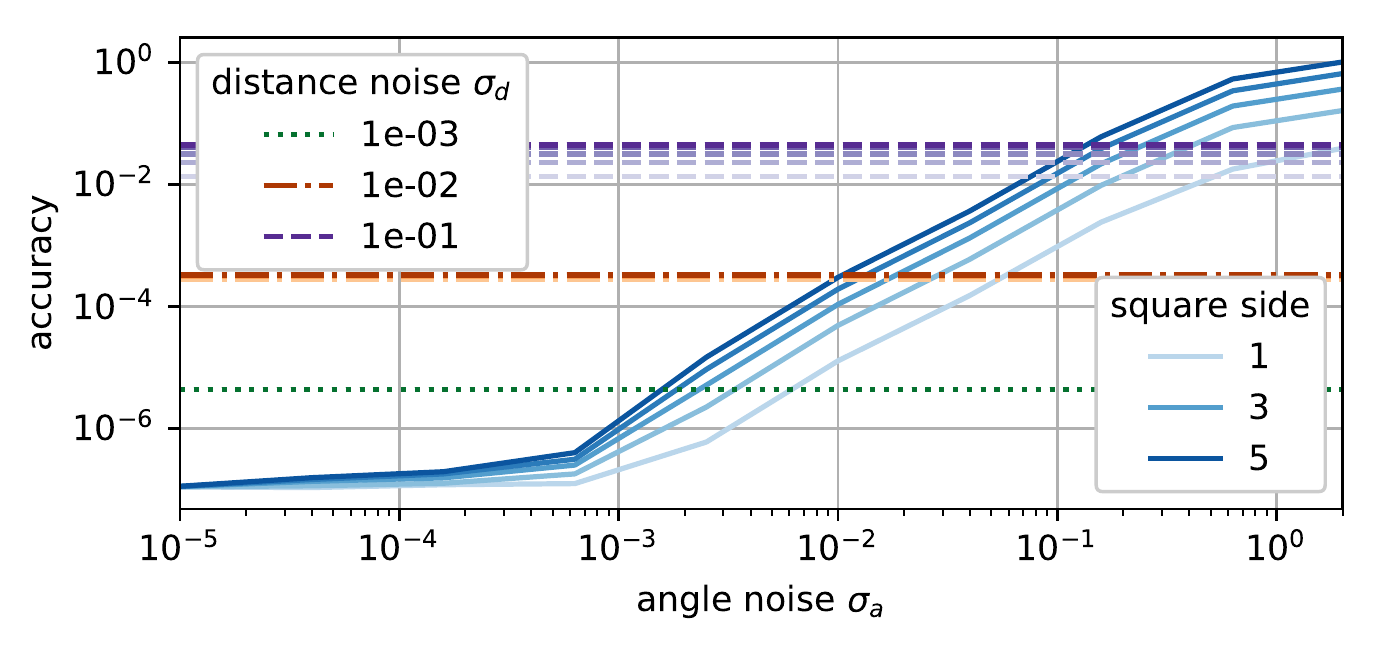}
    \vspace{-2.7em}
    \caption{Point recovery accuracy of 5 points using either distances (dashed lines) or angles (solid lines). The shading corresponds to different sizes of the experiment setup.}%, and $\sigma_a$, $\sigma_d$ denote the standard deviation of angle and distance noise, respectively.}
    \label{fig:angle_distance}
\end{figure}

We also evaluate the performance of the proposed recovery algorithm in terms of point reconstruction \textit{accuracy}, and report results in the second row of Figure \ref{fig:discrepancy}. To this end, we fit the obtained point set with orthogonal Procrustes \cite{Procrustes} to the true locations (including rescaling), and calculate the MSE. We observe that the reconstruction accuracy increases with the number of constraints. Indeed, the plot in the bottom of Figure \ref{fig:discrepancy}, shows a clear correlation between the number of added constraints and reconstruction accuracy. 

Finally, we provide insight into when angle measurements are preferable over distances. For the distance-based reconstruction, we implement the standard multi-dimensional scaling algorithm \cite{EDM2015}. We fix $N=5$ and assume three different noise levels on distances, denoted by $\sigma_d$, and plot the average accuracy over 20 random realizations for a range of angle noise $\sigma_a$. As angle-based recovery is sensitive to the scale of the setup, we show results for different area sizes. Figure \ref{fig:angle_distance} confirms that accuracy decreases with area size and noise, but that for a large regime of angle noise, angle-based localization is more accurate than distance-based localization.

%% file: 7_conclusion.tex
\section{Conclusion and future work} \label{sec:conclusion}

We have proposed necessary and sufficient constraints for the realizability of inner-angle measurements. The constraints are related to well-known trigonometric properties and we show how to construct them both analytically and numerically, given minimal prior knowledge of the point set's topology. We exploit the conditions for a denoising and recovery algorithm and show regimes in which angle-based recovery is preferred over distance-based recovery. 

The required prior knowledge of an equivalency class can be constraining in certain situations, therefore future work will focus on establishing end-to-end algorithms which exploit different levels of constraints in a hierarchical fashion to recover the equivalency class along the way. Partial knowledge of anchor locations could also be exploited to alleviate this problem. Another natural extension of the present work is the incorporation of the realizability conditions in novel methods for outlier rejection, completion, and labeling.